\documentclass[11pt,a4paper]{article}
\usepackage[hyperref]{acl2019}
\usepackage{times}
\usepackage{latexsym}
\usepackage{graphicx}

\usepackage{url}

\aclfinalcopy 

\title{CMU GetGoing: An Understandable and Memorable Dialog System for Seniors}

\author{Shikib Mehri, Alan W Black and Maxine Eskenazi \\
  Dialog Research Center, Language Technologies Institute \\
  Carnegie Mellon University, USA \\
  \texttt{\{amehri,awb,max\}@cs.cmu.edu}}

\date{}

\begin{document}
\maketitle
\begin{abstract}
Voice-based technologies are typically developed for the \textit{average user}, and thus generally not tailored to the specific needs of any subgroup of the population, like seniors. This paper presents CMU GetGoing, an accessible trip planning dialog system designed for senior users. The GetGoing system design is described in detail, with particular attention to the senior-tailored features. A user study is presented, demonstrating that the senior-tailored features significantly improve comprehension and retention of information. 
\end{abstract}

\section{Introduction}

Voice-based technologies have become increasingly common, allowing many end users to benefit from convenient access to information and automated services. Despite this popularity, most systems concentrate on improving the quality of the interaction for the \textit{average user} and are not tailored to the needs of specific sub-populations, for example, seniors. This makes it difficult for older individuals to successfully use the technology. CMU GetGoing has, on the other hand,  specifically been developed as an accessible trip planning dialog system for seniors. 

Seniors may have several unique challenges. As we age, we process information less quickly \citep{diamond2000information}. This suggests that slower information delivery may improve comprehension and information retention. \citet{gordon2005hearing} showed that seniors have difficulty understanding content at faster rates of speech. They particularly struggle with synthesized speech \citep{eskenazi2001study}. As we age we are also more susceptible to external distractions \citep{weeks2014disruptive}, particularly while multi-tasking \citep{clapp2011deficit} (e.g., writing down directions). These issues are a consequence of aging, rather than an unfamiliarity with technology, and as such will continue to be a problem for future generations of seniors. 

These issues were tackled during GetGoing's design and development. To address the slower rate of information processing, GetGoing provides step-by-step information and confirms user understanding. In order to improve comprehension, pauses are inserted to slow the rate of speech. To reduce the impact of distractions, an \textit{attention-grabbing prefix} is inserted prior to delivering any important information. The prefix is similar to the ``summons'' used in \citet{nicolich2016initiations}. Furthermore, since older users are typically less familiar with voice-based assistants, GetGoing relies on several user-directed features: barge-in, flexibility in the order of dialog turns, and allowing the user to correct the system in a conversational manner. GetGoing is presently being used for Southwestern Pennsylvania, but can easily be extended to other geographic regions. The system is delivered over the phone since studies have shown that seniors own fewer smartphones and many have less access to the internet \citep{PEW}.

The paper describes the GetGoing system architecture and its senior-friendly features. An evaluation demonstrates that these features make the information delivered by the system more understandable and memorable for senior users. By identifying and addressing several senior challenges, GetGoing provides seniors with a way to gather relevant transportation information. GetGoing also helps to encourage mobility among seniors by making them more aware of their options when traveling in their city. 

\section{Related Work}

CMU GetGoing is inspired by CMU Let's Go \citep{raux2005let}, which was built for a similar application. GetGoing is an adaptation of this trip planning system, that improves user experience by leveraging recent improvements in speech recognition, speech synthesis, natural language understanding and navigation. Let's Go explored some approaches for improving dialog systems for seniors and non-native speakers \citep{raux2003let}. GetGoing expands on this work by explicitly addressing issues that seniors face. GetGoing also provides driving directions and goes beyond simple directions between specific bus stops.

\citet{langner2005using} explored speaking style modifications to improve understanding of synthesized speech for elderly listeners. \citet{wolters2007making} built on this work, finding that careful word choice and the use of pauses improves senior understanding of synthesized speech. \citet{georgila2010match} presented the MATCH corpus, which is a collection of user-system dialogs for both younger and older users. This corpus was then used by \citet{georgila2010learning} to learn dialog policies using simulated younger and older users. \citet{portet2013design} experimented with seniors' use of voice-based assistants with a Wizard-of-Oz setup and found that older users significantly preferred voice interaction over a tactile system (e.g., a tablet or other touchscreen device). \citet{razavi2019dialogue} created an adaptable dialog management strategy for user-friendly open-domain senior-oriented dialog agents. 

\section{Architecture}

\begin{figure}[h]
    \includegraphics[width=\linewidth]{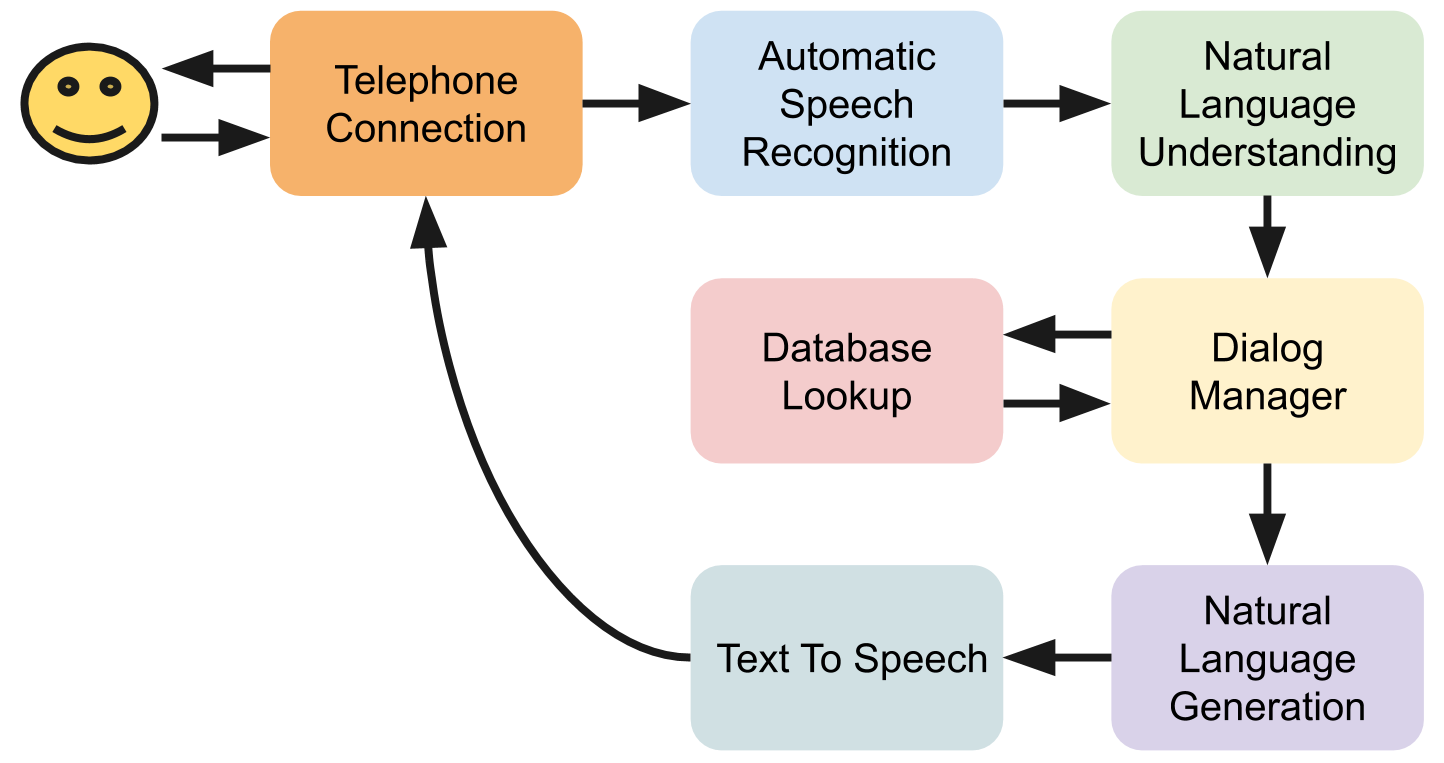}
    \caption{A high-level diagram of the GetGoing dialog system's classic pipeline architecture.}
    \label{dialvis}
\end{figure}

This section describes the architecture of the GetGoing dialog system (Figure \ref{dialvis}). It consists of several standard components of traditional pipeline dialog systems, including automatic speech recognition (ASR), natural language understanding (NLU), a dialog manager (DM), natural language generation (NLG) and text to speech (TTS). The system also consists of a database lookup module, as well as a telephone connection which acts as the user interface. Some architecture components that are not specific to a senior-tailored system are briefly described.

\subsection{Telephone Connection}

The telephone connection is implemented with the Nexmo API\footnote{Nexmo. https://www.nexmo.com/}. This connection can deal with multiple simultaneous calls and has caller identification. The audio signal is streamed from the telephone connection to the ASR. The caller's phone number is encrypted and saved along with the conversation log, to allow for future personalization.

\subsection{Automatic Speech Recognition}

The Google Cloud Speech Recognition API \citep{chiu2018} receives a streamed audio signal and transcribes it in real-time.

Conversations with GetGoing are expected to specifically concern trip planning. As such, an open-vocabulary ASR module could struggle with domain-specific utterances (e.g., street names: "Forbes and Murray"). To this end, we constructed a set of n-grams that are likely to occur in conversations with GetGoing and provided them as ``transcription hints" to the Google Cloud Speech Recognition API. The speech recognition module will then be biased toward this set of n-grams during decoding \citep{aleksic2015bringing}. This set of n-grams is constructed to include several generic responses (e.g., ``Yes", ``No", ``Continue", ``Repeat"...etc.) as well as a list of all street intersections in Pittsburgh (e.g., ``Forbes and Murray'', ``Beechwood and Northumberland'').

The system leverages the real-time interaction with the ASR to implement barge-in \citep{zhao2015incremental}. If the ASR detects a word (speech) in the audio signal, an interrupt signal is sent to the telephone connection and the system output is stopped. The audio signal is also streamed separately and saved.

\subsection{Natural Language Understanding}

The NLU identifies the slot values. 
Slot tagging is formulated as a BIO tagging task \citep{ratinov2009design} wherein a model must assign a label for a sequence of words, corresponding to either a \textit{begin} tag, an \textit{inside} tag or an \textit{other} tag. An example of an utterance labeled in this manner is shown below:

{
\small
\begin{verbatim}
    I               O
    want            O
    to              O
    go              O
    to              O
    Pittsburgh      B-ALOC
    International   I-ALOC
    Airport         I-ALOC
    at              O
    7               B-TIME
    am              I-TIME
    by              O
    transit         B-TRANSIT
\end{verbatim}
}
Given BIO tags for each word in an utterance, the slot values can be obtained by collapsing consecutive words with the same label (e.g., the three ALOC words become \texttt{ALOC:} ``Pittsburgh International Airport''). The BIO tagging model is sophisticated enough to never generate invalid sequences of tags (e.g., an inside token without a corresponding begin token). 

A training set was synthetically constructed for BIO tagging through the use of templates. A total of $55$ templates were manually constructed, including various formulations of valid utterances (e.g., \textit{``I'm leaving from \texttt{DLOC} and going to \texttt{ALOC} at \texttt{TIME}.''}, \textit{``I'm at \texttt{DLOC}.''}). Of these templates, $43$ are different ways of giving arrival/departure/time information, while the remaining $12$ cover simple slot values (e.g., \textit{``Yes'', ``No'', ``Continue'', ``Repeat''}). A set of valid slot values were manually created for each slot type, with examples shown in Table \ref{slots}. These slot values were sampled and inserted into the templates, to construct a total of $19593$ training examples. 

\begin{table*}[]
    \small
    \centering
    \begin{tabular}{|c|c|c|}
    \hline
    Slot Key & Slot Description & Example Slot Values  \\ \hline
    \texttt{DLOC}       &   Departure Location      &   \textit{``CMU''}, \textit{``Smithfield St and Third Ave''}, \textit{``Penn and 26th''}             \\
    \texttt{ALOC}       &   Arrival Location        &   \textit{``Airport''}, \textit{``Baker St at Butler St''}, \textit{``Bayard and Craig''}             \\
    \texttt{TIME}       &   Time                    & \textit{``5 pm''}, \textit{``12:15 pm''}, \textit{``immediately''}, \textit{``right now''} \\
    \texttt{YES}        &   Yes                     & \textit{``Yes''}, \textit{``Sure''}, \textit{``Of course''}, \textit{``Correct''}, \textit{``Fine''}\\
    \texttt{NO}         &   No                      &\textit{``No''}, \textit{``Never''}, \textit{``Negative''}, \textit{``Nah''}, \textit{``Wrong''}\\
    \texttt{PAUSE}      &   Request to pause        &\textit{``Pause''}, \textit{``Wait''}, \textit{``I need a second''}, \textit{``Give me a second''}\\
    \texttt{REPEAT}     &   Request to repeat       &\textit{``Say that again''}, \textit{``Repeat''}, \textit{``One more time.''}, \textit{``What was that?''}\\
    \texttt{CONTINUE}   &   Request to continue     &\textit{``Continue''}, \textit{``Move on''}, \textit{``I'm finished.''}, \textit{``Done''}, \textit{``Next step''}\\
    \texttt{RESTART}    &   Request to restart      &\textit{``Start over''}, \textit{``Begin again''}, \textit{``Restart''}, \textit{``I have another query''}\\
    \texttt{TRANSIT}    &   Transit directions      &\textit{``Bus''}, \textit{``Transit''}, \textit{``I want the bus''}, \textit{``I will transit''}\\
    \texttt{DRIVING}    &   Driving directions      &\textit{``Drive''}, \textit{``By car''}, \textit{``I will take my car''}, \textit{``I want to drive''}\\
    \texttt{CHANGE}     &   Request alternate route &\textit{``Change''}, \textit{``Alternate route''}, \textit{``Different directions''}\\
    \hline
    \end{tabular}
    \caption{A list of the slots used by the NLU.}
    \label{slots}
\end{table*}

A neural mode was trained for BIO tagging. The model is composed of an ELMo embedding layer \citep{peters2018deep} and a bidirectional LSTM \citep{hochreiter1997long}. The ELMo layer provides a generalized latent representation of each word, while the bidirectional LSTM incorporates context and ultimately allows the linear layer to predict the appropriate tag for the word. Since the set of templates and slot values is not exhaustive, a pre-trained generalized embedding layer allows the model to handle unseen grammatical constructs and new slot values.


\subsection{Dialog Manager}

\begin{figure*}[h]
    \includegraphics[width=\linewidth]{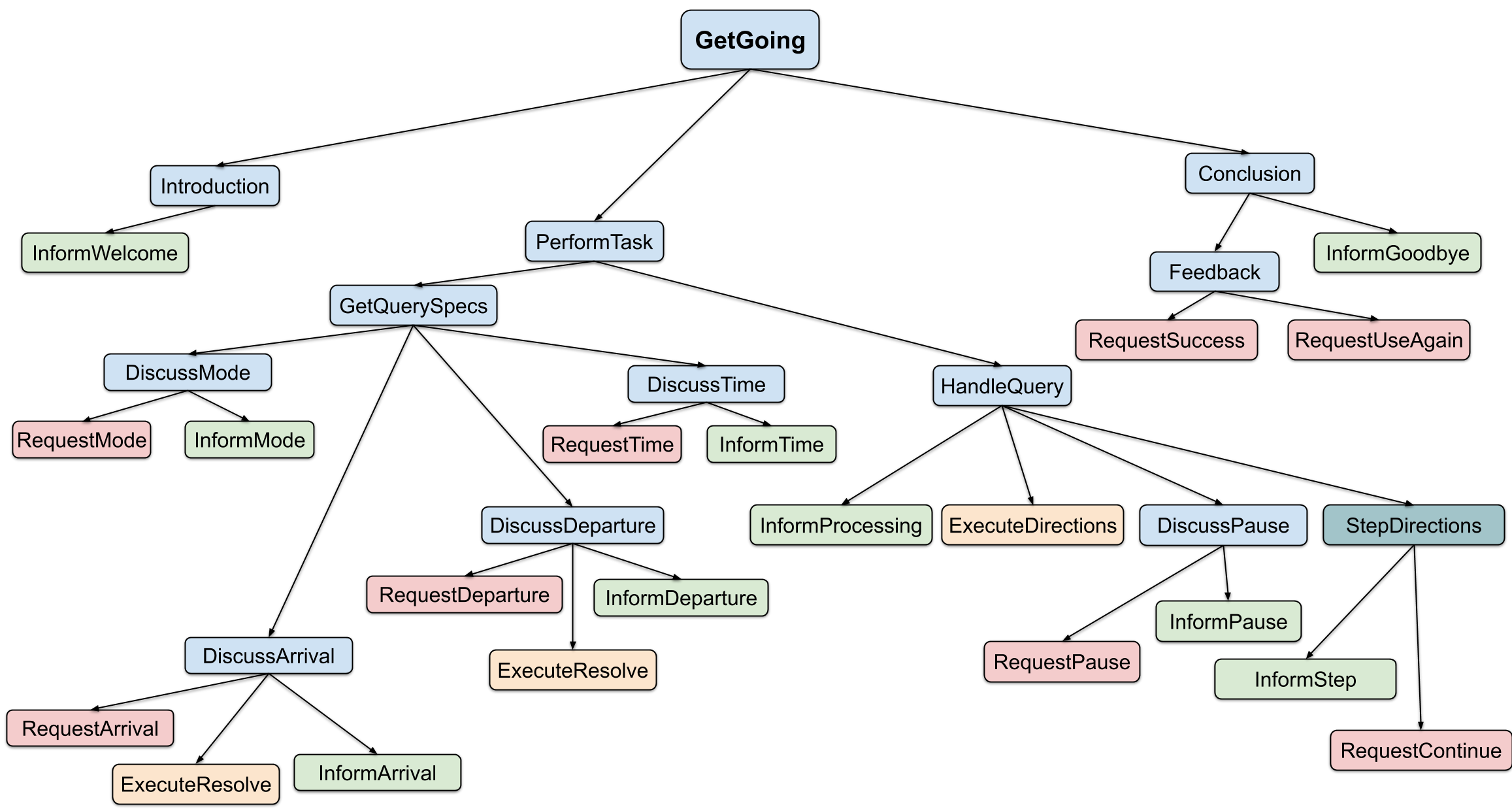}
    \caption{The dialog task tree for GetGoing. Blue nodes are \texttt{DialogAgents}. Green nodes are \texttt{InformAgents}. Red nodes are \texttt{RequestAgents}. Yellow nodes are \texttt{ExecuteAgents}. The turquoise node is a \texttt{StepAgent}. }
    \label{tree}
\end{figure*}

Like Let's Go \citep{raux2005let}, GetGoing's dialog manager (DM) is modelled after the RavenClaw architecture \citep{bohus2009ravenclaw}. The RavenClaw architecture consists of a task-independent dialog engine that carries out a dialog according to a task-specific dialog task tree.  The dialog engine performs an in-order traversal of the dialog task tree, conditioned on the slots output by the NLU. The dialog task tree pictured in Figure \ref{tree}, specifies the various agents in the conversation. \texttt{RequestAgents} requests information from the user and is marked as complete when an appropriate slot value is obtained. \texttt{InformAgents} provides information to the user and \texttt{ExecuteAgents} performs a database lookup and marks it as complete when information is obtained. \texttt{StepAgents} informs the user of a sequence of steps, proceeding to the next step only if the user confirms that they are finished with the current step. \texttt{DialogAgents} do not perform an action themselves, but are instead parent nodes in the task tree.

Following the framework of the RavenClaw architecture \citep{bohus2009ravenclaw}, every node in the task tree contains a set of \textit{concepts} or slots. If the DM is processing a particular node in the tree, the set of active concepts includes those belonging to the current node as well as all of its parents. Given the output from the NLU, the DM can fill any active concepts. For example, the concepts \texttt{DLOC}, \texttt{ALOC}, and \texttt{TIME} belong to the \texttt{GetQuerySpecs} node. As such, when the DM is at the \texttt{RequestArrival} node, it can fill the \texttt{TIME} concept if that information is provided. 

This makes the dialog manager flexible, allowing it to traverse the tree in a user-directed manner rather than going through a set of turns sequentially. The flexibility of the dialog engine allows tasks to be completed out of turn, which is particularly valuable during information elicitation. Users can provide extra information (e.g., \textit{``Where do you want to go?''}, \textit{``I'm going to CMU at 7 PM''}), provide information out of turn (e.g., \textit{``Where do you want to go?''}, \textit{``I'm leaving from the airport''}), or correct the system (e.g., \textit{``Okay, going to CMU. Where are you leaving from?''}, \textit{``No. I'm going to Forbes and Murray''}). This flexibility allows users to interact with GetGoing in mixed-initiative mode, rather than following a rigid system-directed dialog. 

The \texttt{StepsDirection} agent is particularly important for making GetGoing accessible for seniors. By confirming the user's understanding of each step, GetGoing ensures that users have time to comprehend and potentially write down the information. 

\subsection{Database Lookup}

GetGoing uses the Google Maps API\footnote{ https://developers.google.com/maps/documentation/} as its database. The DM queries this database in the three \texttt{ExecuteAgents} shown in Figure \ref{tree}.

The two \texttt{ExecuteResolve} agents occur while the system elicits information about the user's departure and arrival locations. Upon receiving a slot value from the NLU (e.g., \texttt{DLOC:} \textit{``airport''}), the Google Maps API is queried to resolve the text to a valid location (e.g., \textit{``Pittsburgh International Airport''}). This resolving action helps correct potential ASR errors (e.g., \textit{``Beechwood in Northumberland''} to \textit{``Beechwood Boulevard and Northumberland Street''}) and ambiguous locations. The proceeding \texttt{InformAgent} will inform the user of the resolved location that the system understood. 

The \texttt{ExecuteDirections} agent acts after the system has received the complete query. The system queries the Google Maps API to obtain multiple sets of directions between the two locations. The API may return multiple sets of directions. The user can switch between them by asking for an alternate route. Natural language directions are constructed by filling in a template with the information returned by the API. The Google Maps API provides the directions as a sequence of steps. After converting all of the steps to natural language, the steps are passed to the \texttt{StepDirections} agent, which provides the user with the information in an accessible manner.

\subsection{Natural Language Generation}

The natural language generation (NLG) component of GetGoing consists of several hand-written output utterances for every \texttt{InformAgent} and \texttt{RequestAgent} node in the task tree. When these agents are executed by the dialog engine, an utterance is sampled and sent to the TTS. 

The hand-written utterances in the NLG are constructed to ensure that the system is easily understood by seniors. To do this, prior to providing a key piece of information, especially when informing the user of the directions, an \textit{attention-grabbing prefix} is added to each utterance to deflect the user's attention away from other things and toward the ongoing dialog (e.g., \textit{``The next thing you want to do is...''} or \textit{``The final step is to...''}). This also alerts the user that important information is about to be provided, while allowing them time to attention-switch (e.g., from writing things down) or perhaps to tune out distractions before the information is delivered.

 Some portion of the senior population may be unfamiliar with voice-based assistants. As such, system utterances are designed to explicitly provide as much information about using system functions as possible. For example: \textit{``I will pause after every step and give you time to write things down. Feel free to ask me to repeat a step, if you didn't catch it the first time.''}

\subsection{Text To Speech}

Text to speech (TTS) is handled by Nexmo, along with the telephone connection. An SSML string is passed to Nexmo, which synthesizes the speech and streams the speech signal over the telephone connection. This signal can be interrupted for user barge-in as described previously.

Some seniors process information more slowly. To this end, the synthesized speech is "slowed down" by inserting pauses into the SSML prior to speech synthesis \citep{wolters2007making,parlikar2012modeling}. Since the NLG consists of hand-written utterances, pauses have been inserted manually. Long pauses are placed before key pieces of information and at sentence breaks, and shorter pauses are inserted in the middle of complex sentences. Pauses are also inserted to emphasize important information, including bus numbers, street names and departure times. The latter three types of information are provided at increased volume, to further emphasize it.

\subsection{Senior Friendly Design}

CMU GetGoing is built to give seniors information that is more understandable and more easily retained. Following is a list of several relevant features and design decisions. While some of these features may exist in other dialog systems and not be unique to a senior-centric system, together they help make GetGoing information more understandable and memorable.

\begin{enumerate}
    \item \textbf{Confirmation of understanding.} GetGoing confirms that the user understood the system output, and had sufficient time to write it down if they desire. This addresses the concern that older users be able to comprehend and retain instructions.
    \item \textbf{Slower Synthesized Speech.} By inserting pauses prior to synthesizing the speech signal, GetGoing helps to make its utterances more understandable for individuals who process speech more slowly.
    \item \textbf{Attention-Grabbing Prefix.} GetGoing inserts an \textit{attention-grabbing prefix} prior to providing important information. This allows users to switch attention and tune out distractions and focus on the system output.
    \item \textbf{Barge-In.} Allowing users to interrupt the system shifts control of the dialog toward the user. Since senior users may be unfamiliar with voice-based assistants, barge-in allows them to correct system mistakes without having to wait for the system turn to finish. This avoids the user's turns getting "out-of-sync" with the system.
    \item \textbf{Flexible Dialog Manager.} GetGoing's DM is flexible in the order of turns. This allows users to easily correct the system, provide information out of turn, or fill multiple slots at once. This makes the dialog more natural.
    \item \textbf{Telephone Connection.} GetGoing's user interface is the telephone, which reduces the entry barrier for senior users who may not own a smartphone or may be uncomfortable using one.
\end{enumerate}

\section{Evaluation}

This section presents an evaluation of CMU GetGoing. The first subsection evaluates the quality of the NLU model using labeled data. In the second subsection a user study evaluates whether GetGoing is indeed \textit{more understandable} and fosters \textit{better retention} for senior users.  

\subsection{NLU Evaluation}

The GetGoing NLU consists of a neural architecture trained for BIO tagging on synthetic data. The quality of this BIO tagging architecture is evaluated in two settings. First, the accuracy of the model on \textit{unseen slot values} is calculated. Second, the accuracy of the model on \textit{unseen grammatical constructions} is determined. A strong NLU should be able to generalize both to novel slot values and different language formulations. 

The first experiment uses the same synthetic templates, but a different set of slot values. For example, ``Carnegie Mellon University'' is not seen during training, but we still evaluate whether the model can recognize it as a departure location in the right context. The BIO tagging model obtains a label accuracy of \textbf{99.98\%} on the unseen slots, showing that it is able to generalize beyond the slot values that had been seen during training. The second experiment uses a different set of synthetic templates for testing, as well as different slot values. In this second experiment, the model attains an accuracy of \textbf{96.88\%}. The slightly lower score indicates that the model may over-rely on the specific grammatical formulations seen in the training data. This is likely a consequence of using synthetic data. While statistically significant, the performance drop is relatively small and this approach is still appropriate for our use case. 

\subsection{Understandability and Retention Study}

CMU GetGoing is designed to make system output more understandable and memorable for seniors. This user study is designed to quantifiably demonstrate that GetGoing fulfills these goals better than a version of the same system without the senior-specific features. Two variants are compared: \textit{Senior-Tailored Delivery} (SeTD) and \textit{Standard Delivery} (SD).

The SeTD system is the GetGoing system described above. The SD system does not have the three features aimed at improving the accessibility of the system's delivery: (1) confirmation of understanding, (2) slower synthesized speech, and (3) the attention-grabbing prefix. This user study asks the question: \textit{``Do the three GetGoing features enable senior users to better comprehend and retain information?''}

\subsubsection{Study Design}

\begin{table*}[]
\centering
\begin{tabular}{|l|ccc|}
\hline
\textbf{}              & \textbf{SD} & \textbf{SeTD} & \textbf{$p$-value} \\ \hline
\multicolumn{4}{|c|}{\textbf{Younger Users: }Age 18 - 45} \\ \hline
Number of Buses                                 & \textbf{67.65\%}              & 23.08\%                         &  0.0001             \\
Trip Length                                     & \textbf{61.76\%}              & 20.51\%                         &  0.0002           \\
Bus Arrival Times                               & 8.82\%                & 7.69\%                                  &  0.8611            \\
Bus Numbers                                     & 37.67\%              & \textbf{74.36\%}                         &  0.0007           \\
Intermediate Trip Locations                     & 14.71\%                & 25.64\%                                &  0.2484            \\
Average                                         & 32.05\%                & 30.51\%                                &  0.8049            \\ \hline
\multicolumn{4}{|c|}{\textbf{Older Users: }Ages 45+} \\ \hline
Number of Buses                                 & 51.11\%              & 37.21\%                            &  0.1886             \\
Trip Length                                     & 33.33\%               & 25.58\%                           &  0.4261           \\
Bus Arrival Times                               & 6.67\%                & 9.30\%                            &  0.6486            \\
Bus Numbers                                    & 33.33\%              & \textbf{74.42\%   }                         &  0.0001           \\
Intermediate Trip Locations                     & 6.67\%                & \textbf{20.93\% }                          &  0.0495            \\
Average                                         & 26.22\%                & 33.49\%                                &  0.1235            \\ \hline
\end{tabular}
\caption{Slot types provided in responses to question 1, across the two systems. Boldface indicates statistical significance of $p < 0.05$ by t-test.}
\label{study_results}
\end{table*}

The study was given to two different populations: older Amazon Mechanical Turk (AMT) workers (over the age of 45) and younger workers (18 - 45).  All workers had a greater than 95\% HIT approval rate and were paid \$2.00 USD for a task of around 5 - 7 minutes. Each worker completed a survey, which consists of four sections. We note that each participant interacted with only one of the two versions of the system, chosen pseudo-randomly.

The first section included questions about English proficiency and age group. 50 is usually the cutoff age for the seniors, but since AMT has a 45-55 age range, we used 45 as the low part of this range.

Next the workers got their instructions. They were told to call the system, imagining that they were a real user needing directions. The instructions for this are: \textit{``You should imagine that you are a real user of the system, and that you will need to follow these directions soon. This means that you should try to ensure that the system understands what you request and that you understand and remember the directions that it gives you. As a real user would, please do not stop until you get directions.''}

They were then instructed to select any departure and arrival location from a set of six places in Pittsburgh. Prior to making the call, they wrote down which two locations they selected. They were then provided with the GetGoing system phone number and told to make the call. Each worker was either connected to SeTD or SD pseudo-randomly. GetGoing told them that it was either the \textit{orange} system (SeTD) or the \textit{blue} system (SD). The system then asked the caller to immediately enter the specific system they were using. 

After completing the call, workers answered four questions. 

\textbf{Question 1:} \textit{Tell us what you remember about the directions provided by the system? For example, what buses do you need to take?} This question objectively assesses understanding and retention via the task of writing down exact bus numbers and places as they remembered them. Being open-ended, it allowed the user to provide as much information as possible. This question came immediately after the call was over so that the information could still be in recent memory.


\textbf{Question 2:} \textit{Did the system give you useful directions?} The worker was asked their opinion of whether the system had provided useful directions. 

\textbf{Question 3:} \textit{Did you have difficulty understanding the directions provided by the system?} While the first question tests whether the worker understood the system, this self-assessment question asks the worker to judge their understanding. Workers chose between four answers: (1) no difficulty, (2) some difficulty, (3) a lot of difficulty, but could understand and (4) a lot of difficulty and could not understand.

\textbf{Question 4:} \textit{Did you have difficulty providing information to the system?} This question asks the worker to judge how well the system understood them.

After answering these four questions, the workers were directed to the fourth section where they could leave open-ended comments. 

\subsubsection{Study Results}

\begin{table*}[]
\centering
\begin{tabular}{|l|ccc|}
\hline
\textbf{}              & \textbf{SD} & \textbf{SeTD} & \textbf{$p$-value} \\ \hline
\multicolumn{4}{|c|}{\textbf{Younger Users: }Age 18 - 45} \\ \hline
\textbf{Question 2: }Useful Directions      & 79.41\%                   & 87.18\%                           & 0.3720 \\ 
\textbf{Question 3: }User Understand System & 3.18                      & 3.05                              & 0.5473 \\ 
\textbf{Question 4: }System Understand User & 2.97                      & 3.36                              & 0.0627 \\ 
\textbf{Open-Ended: }Comments about difficulty & 5.71\%                & 2.56\%                           & 0.4956 \\ \hline
\multicolumn{4}{|c|}{\textbf{Older Users: }Ages 45+} \\ \hline
\textbf{Question 2: }Useful Directions      & 75.56\%                   & 81.40\%                           & 0.5064 \\ 
\textbf{Question 3: }User Understand System & 3.04                      & 3.07                              & 0.9072 \\ 
\textbf{Question 4: }System Understand User & 3.27                      & 3.30                              & 0.8536 \\ 
\textbf{Open-Ended: }Comments about difficulty & 24.44\%                & 11.63\%                           & 0.1178 \\ \hline

\end{tabular}
\caption{Responses to the subjective questions in the survey. For question 2, the rate of yes answers is shown. For questions 3 and 4, the answers are converted to numeric values with 4 being the most positive answer (e.g., ``system understood me perfectly'') with 1 being the most negative answer (e.g., ``system could not understand me''). The last row shows the percentage of open-ended comments that concerned difficulty with the system, such as speed of delivery and the voice.}
\label{study_results2}
\end{table*}

In all, a total of 230 dialogs were collected: 116 dialogs from older workers and 114 from younger ones. A survey response is deemed valid if the user requested public transit directions (not driving), and GetGoing provided them with directions. A total of 17 users had difficulty interacting with the system and getting it to understand their request, primarily due to ASR errors. After filtering out these responses and any work that had been done incorrectly (got driving, not bus directions, for example), there were 88 dialogs from seniors (43 for SeTD and 45 for SD) and 73 from non-seniors (39 for SeTD and 34 for SD). 

Results for the first question are shown on Table \ref{study_results}. The workers gave open-ended responses that were quantified by counting the slot types provided in each response. There are five slot values: number of buses, trip length, bus arrival times, percentage of bus numbers (what percentage of all the bus numbers they were given did they recall on average), and intermediate trip locations. Since GetGoing explicitly states the number of buses and the trip length at the very beginning of the directions, these two slots were the easiest to remember \citep{jahnke1965primacy}. The SD workers retained this information better than the SeTD workers. Bus numbers and intermediate trip locations require the user to remember unfamiliar entity names, making them more difficult to understand. Also, since they occurred within the flow of directions, not at the beginning nor the very end, they were more difficult to isolate and remember. These two slots were retained better by the SeTD workers. Examples of answers to this question are provided below, the first two examples are from workers who spoke to SD and the third from a worker who spoke to SeTD:

\begin{itemize}
    \item ``2 busses...one was 61C, the other was not clear''
    \item ``I will only need to take one bus, 28X. It leaves at 10:19 AM and my trip will be 1 hour 13 minutes. I will arrive at the lower level of the airport.''
    \item ``28x to Forbes, opposite Bellefield depart 3:10pm. 61D to Target drive, depart 4:15 pm. Walk to 149 West Bridge Street''
\end{itemize}

Younger users retained information well for both systems. Those who spoke to SD tended to remember the number of buses and the trip length best, while those who spoke to SeTD retained the bus numbers best. On the other hand, older users retained more information when speaking to SeTD with a statistically significant difference in the retention of bus numbers and trip locations. For older users, the average number of slots retained improved greatly (\textbf{+7.27\%}) when using SeTD while it slightly decreased for younger users (\textbf{-1.54\%}). It is important to note that younger users perform reasonably well with SD, proving that while SD is not tailored to seniors, it is a good, understandable system. These results show that the features added to GetGoing enable older users to understand and retain the information provided by the system.

The answers for the three other questions are shown on Table \ref{study_results2}. For older users there is a slight, but insignificant, improvement over the baseline system across the questions which required subjective self-assessment. Since workers only spoke to one system, they had to assess their interaction without a frame of reference. The responses to Question 3 indicate that workers \textit{believed} that they effectively understood both systems, however, when compared to the results of question 1, those who spoke to SeTD understood much more than they said they thought they had on Question 3. Analysis of the seniors' responses shows a weak correlation between the subjective answers to Question 3 and the objective answers to Question 1 (specifically, the bus number slot) with a Spearman coefficient of 0.15 and a Pearson coefficient of 0.18. This suggests that objective questions are better indicators of user experience. Subjective measures appear to be insufficient when workers don't have a suitable point of comparison.

The last question asked workers to leave open-ended comments. It should be noted that several older users of SD complained about the speed and delivery of the directions. Out of the 45 senior responses for SD, 11 workers left unprompted feedback that the directions were given too fast. Out of the 43 responses for SeTD, only 5 workers left similar feedback. This implies that the SeTD features offset system delivery issues. While senior users left comments about SD 24.44\% of the time, younger users only left comments 5.71\% of the time. This difference is statistically significant with $p=0.0259$, suggesting that speed of delivery is of particular concern to older users.  

\section{Conclusion}

CMU GetGoing is a trip planning dialog system for seniors. This paper presents the architecture of GetGoing, with particular attention to features that are designed to help older users better understand what the system is saying. A user study  shows that GetGoing's senior-tailored delivery improves comprehension and retention of information. In the future, the system will be improved further by concentrating on improving system understanding of older users and users' comprehension of system directions. Specifically future work will in part concentrate on improving the system's delivery to increase the retention of bus arrival times and intermediate trip locations.

\section{Acknowledgements}

This work was supported by CMU Mobility21 from the US Department of Transportation. We thank the workers on Amazon Mechanical Turk for making our research possible.

\bibliography{acl2019}
\bibliographystyle{acl_natbib}
\end{document}